\documentclass{amia}
\pdfoutput=1
\usepackage{todonotes}
\usepackage{times}
\usepackage{latexsym}
\usepackage{graphicx}
\usepackage[T1]{fontenc}
\usepackage{microtype}
\usepackage{booktabs}
\usepackage{url}
\usepackage{amsmath}
\usepackage{xcolor}
\usepackage[font=small,labelfont=bf]{caption}
\usepackage{subcaption}

\usepackage{array}
\usepackage{makecell}
\usepackage{multirow}

\usepackage{float}

\begin{document}

\title{Entity Anchored ICD Coding}
\author{Jay DeYoung, MS$^1$\footnotemark[1], Han-Chin Shing, PhD$^2$\footnotemark[1], Luyang Kong, MS$^2$, \\Christopher Winestock, MD PhD$^2$, Chaitanya Shivade, PhD$^2$}
\institutes{$^1$ Northeastern University, Boston, MA \\ $^2$Amazon, Seattle, WA \\ 
}

\renewcommand{\thefootnote}{\fnsymbol{footnote}}
\footnotetext[1]{equal contribution}
\footnotetext[2]{Accepted to American Medical Informatics Association (AMIA) 2022 Annual Symposium}


\maketitle



\renewcommand{\thefootnote}{\arabic{footnote}}

\section*{Abstract}

\hspace{-0.6em}
\textit{
Medical coding is a complex task, requiring assignment of a subset of over 72,000 ICD codes to a patient's notes. 
Modern natural language processing approaches to these tasks have been challenged by the length of the input and size of the output space.
We limit our model inputs to a small window around medical entities found in our documents.
From those local contexts, we build contextualized representations of both ICD codes and entities, and aggregate over these representations to form  document-level predictions.
In contrast to existing methods which use a representation fixed either in size or by codes seen in training, we represent ICD codes by encoding the code description with local context.
We discuss metrics appropriate to deploying coding systems in practice. We show that our approach is superior to existing methods in both standard and deployable measures, including performance on rare and unseen codes.
}


\section{Introduction} \label{sec:intro}

Medical coding is the complex task aimed at generating a coded summary of clinical information associated with a patient. It is typically carried out by trained professionals who examine relevant parts of the patient's Electronic Health Record (EHR) to assign healthcare terminology codes. 
The International Classification of Diseases (ICD) is a global tool, endorsed by the World Health Organization (WHO), for describing medical diagnoses and procedures. 
Hospitals, Doctors, and countries worldwide use ICD for reporting mortality, morbidity, patient procedures, and insurance claims.  In some countries, particularly the United States of America, ICD coding is the primary method by which hospitals and insurance companies identify service fees, define procedures, and negotiate reimbursement and billing. 


We focus on the 10th revision of ICD (ICD10), specifically the ICD10 Clinical Modifications (CM). ICD10-CM can be viewed as a hierarchical ontology consisting of over 90,000 codes. In the ICD-ontology inner nodes are general descriptors of medical conditions (e.g. the ICD10-CM code S62.5 corresponds to \textit{Fracture of thumb}) and leaf nodes contain very specific information from the parent (e.g. S62.501B corresponds to \textit{Fracture of unspecified phalanx of right thumb...}). Billing regulations require claim information to be as specific as possible; one of the 72,000 leaf nodes. 


Medical claim processing in the Unites States involves a healthcare provider 
compiling coded care information and transmitting it to a payor 
typically via a third party clearing house. The payer validates the claim for accuracy and compliance and determines an amount the provider will receive,
 outstanding balances charged by the patient.
A coded claim form shared by all parties has to follow a predefined format and regulations.
Common to these forms is a ranked list of billable ICD10-CM diagnoses codes and a set of procedure codes specifying the treatment for those diagnoses. 


As this is expensive and time consuming, researchers have long explored computer assisted methods for medical coding \cite{Stanfill2010}. 
Early work focused on rule-based systems \cite{Farkas2008AutomaticCO}. 
Later researchers explored using traditional machine learning techniques to address the problem \cite{Kavuluru2015}. 
More recently, deep learning methods 
(Section \ref{sec:related_work}) have demonstrated significant improvements. 
However, a large body of work models is aimed at directly generating billable codes from input text; most models do not attempt to report a cause (e.g. a medical entity) for a particular coding prediction.
While evaluating the task based on traditional machine learning based metrics for classification is reasonable, an evaluation reflecting real-world application is missing. 
Finally, transformer based models have shown significant improvements across several NLP tasks \cite{wang2018glue,wang2019superglue,Nye2020UnderstandingCT,Gu2022DomainSpecificLM}. 
With a few notable exceptions \cite{Dong2021}, transformer based models have not been studied in detail for the task of medical coding.


In this paper, we focus on the problem of automatically extracting ICD10-CM codes from a clinical note and ranking them for the purpose of claim processing. The goal is to develop a system that can assist a professional perform medical coding. Our contributions are three-fold: a grounded transformer model that includes evidence for its predictions; successful scaling and integration of transformer models into clinical coding; and a discussion and analysis of metrics suitable for coding systems in practice.








\section{Related Work} \label{sec:related_work}


Larkey and Croft\cite{larkey1996combining} is one of the earliest work to explore assigning ICD codes to inpatient discharge summaries. They demonstrate that combining three classifiers (K-nearest neighbors, relevance feedback, Bayesian independence) yields higher performance than the individual components.
Other early work\cite{Lima1998AHA} leverages cosine similarity between discharge summaries and ICD codes to build a hierarchical classifier. 
Later, traditional machine learning models such as support vector machines  \cite{farkas2008automatic, Kavuluru2015} were also applied to investigate the problem. 

Recent years have also seen the application of deep learning methods to the task of medical coding~\cite{shi2017toward,xie2018neural,shing2019assigning}. The most notable work among them is Convolution Attention for Multi-Label (CAML) classification \cite{Mullenbach2018}. CAML uses an attention mechanism to aggregate information across a document encoded with convolutional neural nets in a multi-label classification setting. The CAML paper reports performance on the full set, and the top-50 ICD9 codes. Experiments are conducted on discharge summaries from the MIMIC-III dataset~\cite{Johnson2016MIMICIIIAF}. The tokens corresponding to attention weights are used as a tool for interpreting the predicted ICD9 codes. CAML still serves as a strong model in ICD9 coding today.

The advent of transformer models \cite{vaswani_attention,devlin-etal-2019-bert} has revolutionized natural language processing, and they have started making their way into clinical coding. 
The transformer model fundamentally relies upon self-attention \cite{vaswani_attention} to form document-level contextualized word representations. 
BERT \cite{devlin-etal-2019-bert} was the first large scale transformer model. 
It was trained as a denoising auto-encoder on general-domain text - that is to predict a (potentially missing or scrambled word) from contextual clues. 
More recently, biomedical versions of these transformer language models have been released \cite{alsentzer-etal-2019-publicly,Huang2019ClinicalBERTMC,Lee2020BioBERTAP}. 

Most works involving transformers for medical coding approach the task as a multi-label classification problem \cite{Amin2019MLTDFKIAC,Zhang2020BERTXMLLS,Ji2021DoesTM,Pascual2021TowardsBA,Feucht2021DescriptionbasedLA}, often borrowing from CAML. 
These models vary in approach for handling document lengths; some attempt a hierarchical approach\cite{Amin2019MLTDFKIAC,Feucht2021DescriptionbasedLA}, some expand the underlying language model itself \cite{Zhang2020BERTXMLLS,Pascual2021TowardsBA,Feucht2021DescriptionbasedLA}, and some truncate or encode a document in sections\cite{Ji2021DoesTM,Pascual2021TowardsBA}. All papers report difficulty integrating these approaches, often underperforming existing methods.


Additionally, these systems primarily report classification metrics, ignoring ranking metrics or proxies for evaluating real-world usage. 
Most systems such as CAML and DLAC\cite{Feucht2021DescriptionbasedLA} report results on the top 5 or 8 retrieved codes.
Additionally, interpretability is a secondary concern with most, and systems producing scores use derived metrics (e.g., model attention) now known to be misleading \cite{Jain2019AttentionIN,Wiegreffe2019AttentionIN} for providing evidence for model results.

\section{Data} \label{sec:data}

\begin{table}[b    ]
\small
\centering
\begin{tabular}{@{}lrrrrr@{}}
\toprule
          & \# document & \# code         & \# entity       & \# words   & \# sentences \\ \midrule
MIMIC-III & 186/28/22   & 6,924/940/833    & 12,299/1,708/1,438 & 1,525$\pm$1,045 & 87$\pm$59       \\
Internal      & 381/53/41   & 10,822/1,554/1,167 & 17,728/2,595/1,969 & 941$\pm$637   & 63$\pm$54       \\ \bottomrule
\end{tabular}
\caption{Dataset statistics for MIMIC-III and Internal in \textbf{train}/\textbf{validation}/\textbf{test} partitions. "\# document" is the number of documents in each dataset. "\# code" is the sum of unique codes in each document. "\# entity" is the number of entities, whereby multiple entities can link to the same code. The number of words and sentences are represented as \textbf{mean}$\pm$\textbf{std}. \label{tabl:stats}}
\end{table}

\begin{table}[t]
\small
\centering
\begin{tabular}{@{}lrrrrr@{}}
\toprule
                 & \multicolumn{1}{l}{Train} & \multicolumn{2}{l}{Validation} & \multicolumn{2}{l}{Test}     \\ \midrule
                 & \# unique code            & \# unique code  & \# mismatch  & \# unique code & \# mismatch \\ \midrule
MIMIC III        & 618                       & 127             & 45 (35.4\%)  & 112            & 44 (39.3\%) \\
Internal             & 829                       & 216             & 53 (24.5\%)  & 190            & 52 (27.4\%) \\
MIMIC III + Internal & 1204                      & 302             & 74 (24.5\%)  & 273            & 80 (29.3\%) \\ \bottomrule
\end{tabular}
\caption{Unique ICD10-CM codes for MIMIC-III and Internal, by data subset. The ICD10-CM code distribution is particularly long-tailed; many codes in validation and test sets cannot be found in the training set. "\# mismatch" is the number of out of domain codes, and we show the percentage of unique codes not found in validation and test set.}
\label{tabl:code_mismatch}
\end{table}

\begin{table}[t]
\vspace{1em}
\small
\centering
\resizebox{\textwidth}{!}{\begin{tabular}{@{}lll@{}}
\toprule
\multicolumn{3}{l}{\hspace{-0.8em} \textbf{Example Document}} \\ \midrule
\multicolumn{3}{l}{\begin{tabular}[c]{@{}l@{}}Admission Date :  2300-2-30  \qquad \qquad \qquad \qquad \qquad \qquad \qquad \qquad \qquad \qquad \qquad \qquad \qquad \qquad \quad Discharge Date :  2300 - 3 - 2 \\ Date of Birth :   2293-1-6  \qquad \qquad \qquad \qquad \qquad \qquad \qquad \qquad \qquad \qquad \qquad \qquad \qquad \qquad \qquad \qquad \qquad \qquad \qquad \quad Sex :  M \\ Service :  Emergency Medical Centers of America Medicine \\ HISTORY OF PRESENT ILLNESS :  Patient is a 7 - year - old male with past medical history significant for \colorbox[HTML]{df65b0}{pica \textbf{F983}}\\  post recent hospitalization and for \colorbox[HTML]{c994c7}{seizures \textbf{G40509}} , who presents after an \colorbox[HTML]{dd1c77}{overdose \textbf{T450X1A}} of Benadryl.  Per report , \\the patient took approximately 10 tablets of  Benadryl on the night prior to admission .  The patient ' s father last saw the patient at \\ approximately 11 p.m. at which time he was doing well .  He found the patient the next morning at 3 a.m. to be \colorbox[HTML]{f1eef6}{lethargic \textbf{R5383}} \\ with \colorbox[HTML]{d4b9da}{nausea \textbf{R112}} and \colorbox[HTML]{d4b9da}{vomiting \textbf{R112}} .\\ \qquad \qquad \qquad \qquad \qquad \qquad \qquad \qquad \qquad \qquad \qquad \qquad \qquad \qquad \qquad \qquad \qquad \qquad \qquad {[}{[}Firstname Lastname{]}{]} , M.D.  1337\\ Dictated By : {[}{[}Firstname Lastname{]}{]}\end{tabular}} \\ \midrule \midrule
\textbf{Ground Truth}    & \textbf{ICD10}     & \textbf{ICD10-CM Code Description}      \\ \midrule 
Primary Code                                                                                                                                                                                                                                                                                   & \colorbox[HTML]{dd1c77}{T450X1A}                                                                                                                                                                                                                                                                               & Poisoning by antiallergic and antiemetic drugs, accidental (unintentional), initial encounter                                                                                                                                                                                                                                                                               \\
Secondary Code                                                                                                                                                                                                                                                                                 & \colorbox[HTML]{df65b0}{F983}                                                                                                                                                                                                                                                                                  & Pica of infancy and childhood                                                                                                                                                                                                                                                                                                              \\
Secondary Code                                                                                                                                                                                                                                                                                 & \colorbox[HTML]{c994c7}{G40509}                                                                                                                                                                                                                                                                                  & Epileptic seizures related to external causes, not intractable, without status epilepticus                                                                                                                                                                                                                                                                                                                                         \\
Irrelevant Code                                                                                                                                                                                                                                                                                & \colorbox[HTML]{d4b9da}{R112}                                                                                                                                                                                                                                                                                  & Nausea with vomiting, unspecified                                                                                                                                                                                                                                                                                                                                   \\
Irrelevant Code                                                                                                                                                                                                                                                                                & \colorbox[HTML]{f1eef6}{R5383}                                                                                                                                                                                                                                                                                 & Other fatigue                                                                                                                                                                                                                                                                                                                                                       \\ \bottomrule
\end{tabular}}
\caption{An example document, modified for patient's privacy protection. Comprehend Medical tags medical entities in the document, and links them to candidate ICD10-CM codes, represented as \colorbox{lightgray}{entity \textbf{ICD10}}. Multiple entities can link to the same code (e.g., R112). Our models take in the candidate ICD10-CM codes, with surrounding context and code descriptions to predict and rank the candidates as the primary code, secondary codes, or irrelevant codes (code that should not be included for billing purposes).}
\label{tabl:example}
\end{table}

We use two English-language clinical datasets: de-identified documents in outpatient settings purchased from a data vendor, and a subset of MIMIC-III's discharge summaries~\cite{Johnson2016MIMICIIIAF}. 
Though MIMIC-III has ICD9 annotations, it does not have ICD10-CM annotations nor entity-level annotations. 
Both of these datasets are thus doubly-annotated for ICD10-CM codes by internal annotators with professional medical coding experience. Annotators independently annotated data and met with a third senior annotator to resolve conflicts. The annotations were collected in three steps: 1. Named entities of type \textit{medical condition} (these include diagnoses, signs, and symptoms) were annotated. 2. For all entities of the type \textit{medical condition}, annotators could search the ICD10-CM ontology for relevant codes and assign a code that they felt was most appropriate. 3. Finally, the ICD10-CM codes were assigned a rank in the range of 1 to a maximum of 12, where 1 was the primary diagnosis, most relevant to billing and 12 being the least relevant. We chose 12 as the upper limit because that is the maximum number of codes a healthcare provider can enter in the billing claim in a professional setting, i.e. form 837P. Note that all ICD10-CM codes linked to medical condition entities that did not have a rank were marked as "not relevant to billing".
Table \ref{tabl:example} contains a short example from MIMIC-III of annotated data. The MIMIC-III subset consists of 236 documents, the internal dataset consists of 475 documents. 
Between these datasets, we have 22,240 total coding instances, ranked for relevance to overall billing (Table \ref{tabl:stats}). 
By unique code count, around 30\% of the codes seen in the validation and test sets are unseen in the training sets (Table \ref{tabl:code_mismatch}).






\section{Models} \label{sec:models}
Most transformer models used in ICD-coding are limited by the maximum length input they can process.
Established methods (Section \ref{sec:related_work}) typically limit or segment the input.
As the inputs in ICD-coding vary greatly in length - some are composed of short notes, while some are substantially longer (Section \ref{sec:data}), these methods can be unreliable.
In contrast, our approach is relatively insensitive to document length, as we work on entities and contexts, which we represent independently and aggregate as needed.
We operate as a pipeline: first we extract  medical entities and candidate ICD10-CM codes, then we rank the ICD10-CM codes.
Our models thus function by taking a medical entity, its context, and candidate ICD10-CM codes, building a representation of the code and contexts, and providing a final relevance decision for each code.
Our models are BERT-based \cite{devlin-etal-2019-bert}, trained on either clinical notes \cite{alsentzer-etal-2019-publicly,Huang2019ClinicalBERTMC} or in the general domain \cite{Beltagy2020Longformer}.

\vspace{-1em}
\paragraph{Entity \& Code Extraction.}
For all models, we begin with outputs from Amazon Comprehend Medical ICD10-CM linking API \footnote{\url{https://docs.aws.amazon.com/comprehend-medical/latest/dev/textanalysis-entitiesv2.html} \url{https://docs.aws.amazon.com/comprehend-medical/latest/dev/ontology-icd10.html}}.
Documents are passed through Comprehend Medical to tag medical named entities.\cite{bhatia2019end} An ICD10-CM ontology linking module is then used to link potentially relevant ICD10-CM codes to these entities.\cite{zhu2020latte, kong2021zero} These linked ICD10-CM codes, referred to as ``candidates'', are then doubly annotated with disagreement resolved by an experienced medical coder (Section~\ref{sec:data}). These annotated candidate ICD10-CM codes are then used as input to our models described below, where they re-rank the candidates to determine the relevance and ranking of the ICD10-CM codes.


\subsection{Ranking}
\label{sec:ranking}

\begin{figure*}[t]
    \centering
    \includegraphics[width=0.7\textwidth]{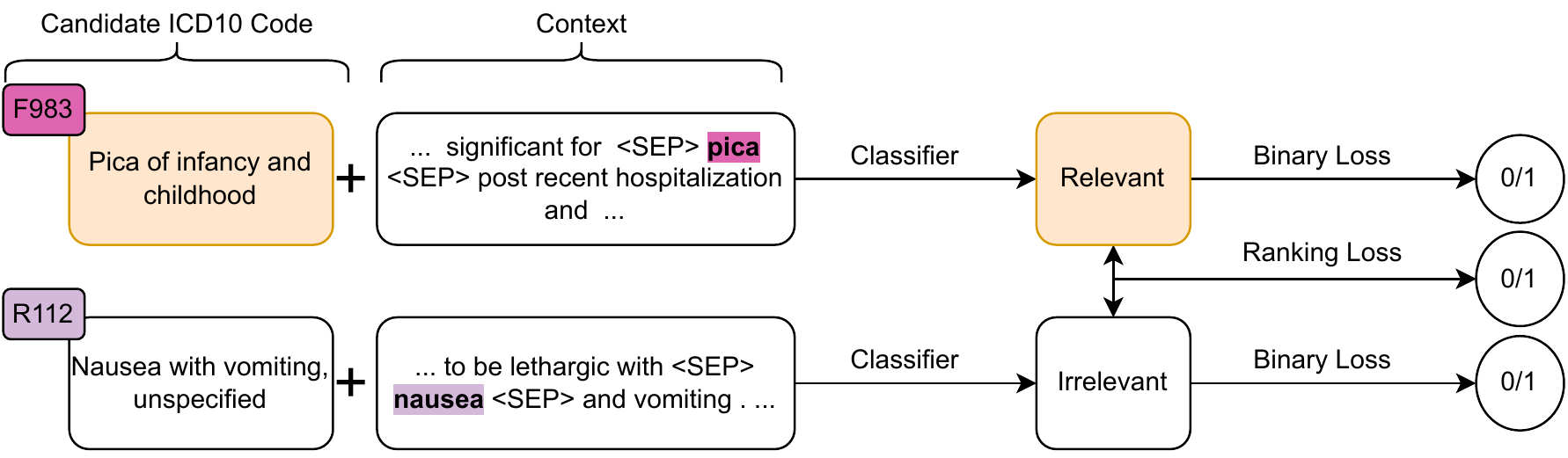}
    \caption{Example Base Models (BERT): for a single instance/ICD code, estimate relevance. Optionally add a loss component for relative instance ranking.}
    \label{fig:simple_bert_model}
\end{figure*}

\begin{figure*}[t]
\centering
\begin{subfigure}{.5\textwidth}
    \centering
    \includegraphics[width=0.95\textwidth]{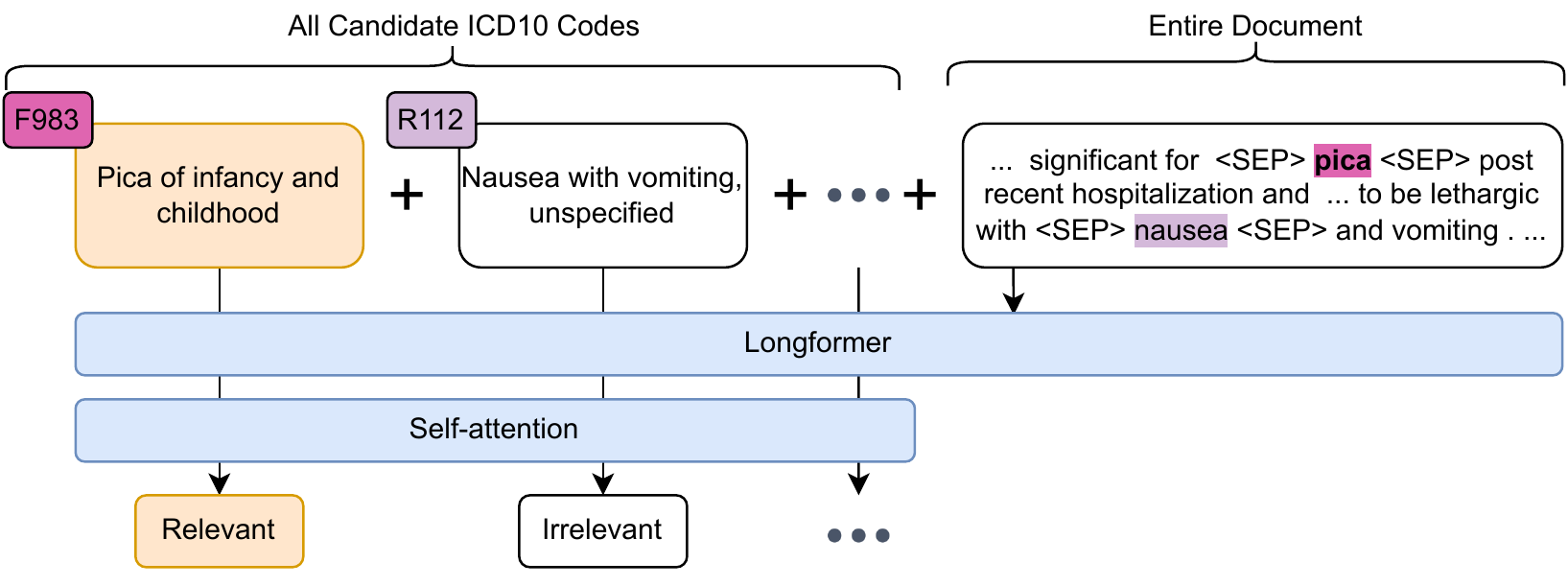}
    \label{fig:longformer_code_based}
\end{subfigure}%
\begin{subfigure}{.5\textwidth}
    \centering
    \includegraphics[width=0.95\textwidth]{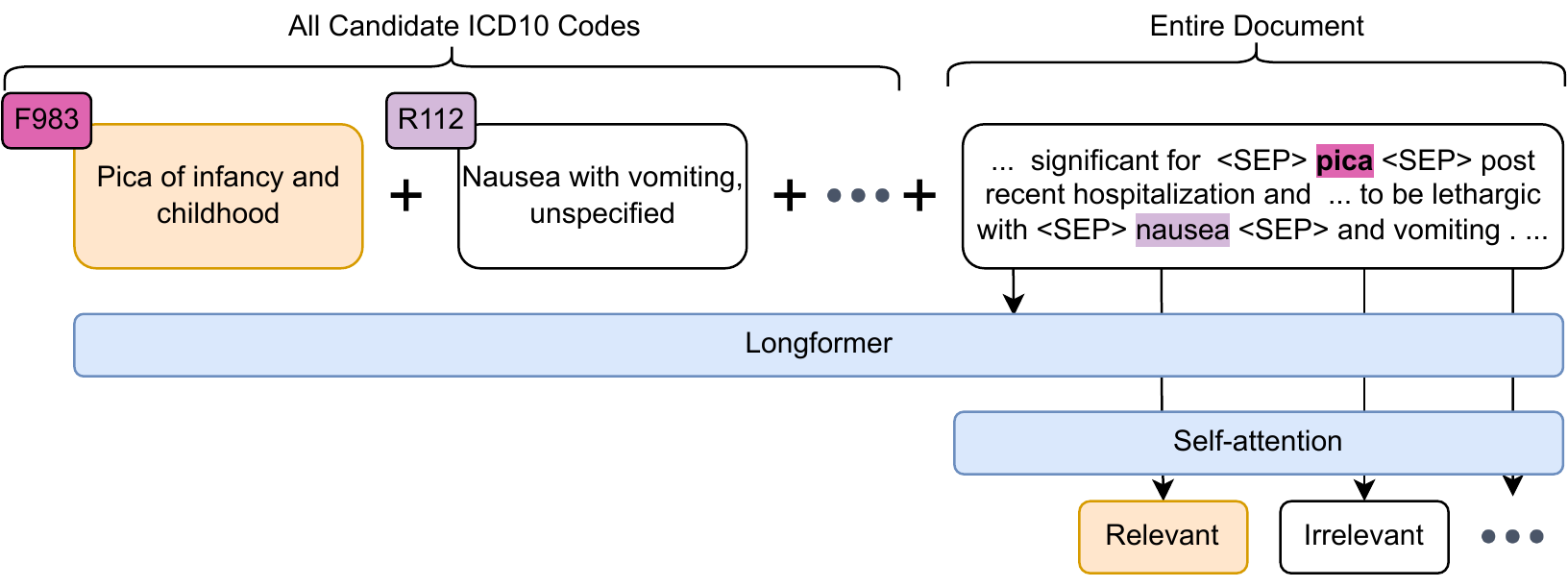}
    \label{fig:longformer_entity_based}
\end{subfigure}
\caption{(\textit{Left}) Code-level relevance assignment. Each ICD10-CM code description (plus a starting CLS token) is concatenated with the document text. The first token of each ICD10-CM code and each entity receives global attention and is encoded with Longformer. The model self-attends over the contextualized representation of each ICD10-CM code (via the CLS token) to classify each code's overall relevance. (\textit{Right}) Entity level relevance assignment. Encoding as in (\textit{left}), using the first entity token instead of the ICD token. The overall entity billing ranking is then used to rank the top-1 retrieved ICD10-CM code for that entity.}
\label{fig:longformer}
\end{figure*}

\textbf{Base Model}. Once we have all the potential codes associated with an input note, we can begin ranking them for relevance. 
Our base model in Figure \ref{fig:simple_bert_model} concatenates an ICD10-CM code's textual label with an entity and context, and produces a unary relevance decision. 
We start with a simple clinical BERT based model \cite{Huang2019ClinicalBERTMC}: we concatenate the ICD10-CM code description with a medical entity and surrounding tokens, and using the \texttt{CLS} token\footnote{a special token to aggregates representations for classification}, predict a unary score using a sigmoid. We denote the entity of interest via \texttt{<sep\_token>}\footnote{a special token designating a text boundary}. We minimize a binary cross-entropy loss: 
$$-\frac{1}{\|\text{docs}\|}\sum_{d \in \text{docs}} \frac{1}{\|c,e\|}\sum_{(c,e)\in d} \left( \text{Rel}_{c,e} \log p(\text{Rel}_{c,e}) + (1-\text{Rel}_{c,e}) \log (1-p(\text{Rel}_{c,e})) \right)$$
Where $d$ is a document, $c$ an ICD-code, $e$ an entity, $\text{Rel}_{c,e}$  an indicator of code and entity relevance in the document ($d$ elided), and $p(\text{Rel}_{c,e})$  the model's prediction of entity/code relevance, in this case a sigmoid over the \texttt{CLS} token.


\textbf{Longformer Models}. Expanding beyond local context, we introduce document level information via Longformer \cite{Beltagy2020Longformer}, a transformer model allowing for for rich interactions between distant elements within a document. 
These interactions take place via a global attention, where some tokens attend to and are attended to by all other tokens, at the expense of a smaller local attention. 
We use this global attention to interact both ICD10-CM code descriptions and medical entities with each other and the entire document (Figure \ref{fig:longformer}). We train these models in two contexts: to directly predict an ICD10-CM code's importance (Figure \ref{fig:longformer} left), and to directly predict an entity's importance (Figure \ref{fig:longformer} right). For both cases, we take the first token (of the entity or the code), and form a linear layer followed by a sigmoid. These models are trained as the base model above. When predicting an entity's importance, we use the code attached to that entity for ranking purposes.

\begin{figure*}[t]
\centering
\includegraphics[width=0.7\textwidth]{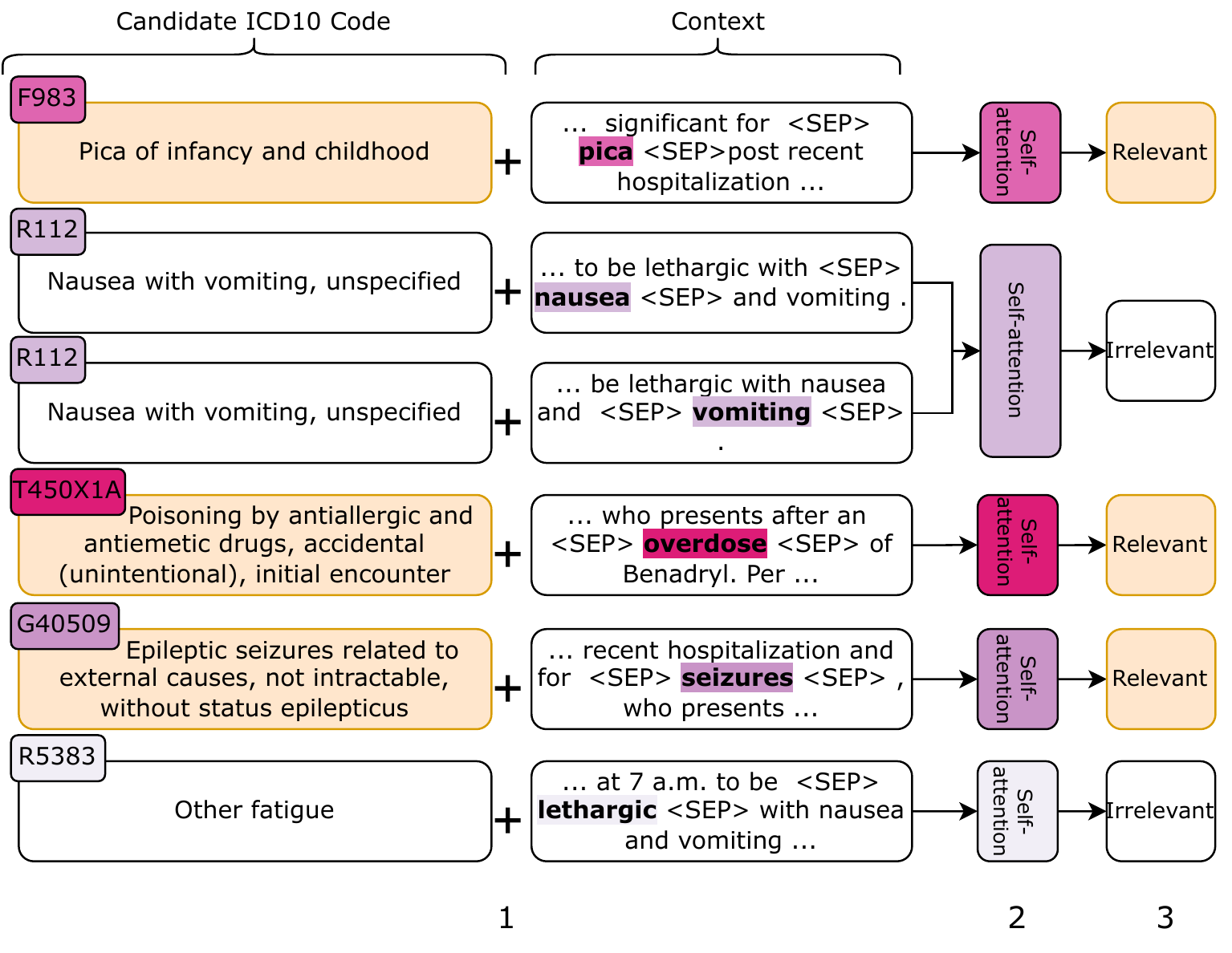}
\vspace{-1em}
\caption{Example of the three-phase aggregation model. (1) Encode as in the base model. (2) For each ICD-code, aggregate the representations using self-attention over the \texttt{CLS} tokens, forming a document-level representation for solely that code. (3) Classify each code as relevant or not, as in the base model.}
\label{fig:retrieval_model}
\end{figure*}

\textbf{Aggregation Model}. A downside of the Longformer models is the co-mingling of context and codes - the model loses the attachment of the code to the local context in which it was retrieved.
As a solution, we return to the base model. 
Rather than classifying relevance  at the entity level, we aggregate across evidence for the code's relevance at local and document level contexts (Figure \ref{fig:retrieval_model}). 
We self-attend to the representation of each code and its evidence, producing a fully contextualized representation for each code. 
We use the same unary prediction method as in the base model.

\textbf{Ranking Loss}. In order to introduce information about relative code importance, we add a \textit{ranking loss} to each model. The ranking loss is designed to force relevant and irrelevant codes apart, to induce differences in codes of different billing relevances. The ranking loss for a document is then:
$$
-\frac{1}{s}\sum_{(c_1,e_1) | \text{Rel}_{c_1,e_1}=1} \sum_{(c_2,e_2) | \text{Rel}_{c_2,e_2}=0}  \left( p(\text{Rel}_{c_1,e_1}) -  p(\text{Rel}_{c_2,e_2}) \right)
$$
where $s=\|(c,e)  | \text{Rel}_{c,e}=1\| \|(c,e) | \text{Rel}_{c,e}=0\|$, the number of pairings in the sum above, and $d$ elided.
BCE-losses and ranking losses are then interpolated, on a per-document basis, with a hyperparameter $\lambda$.



\vspace{-1em}
\paragraph{Baseline: CAML.} Convolutional Attention for Multi-Label classification (CAML)~\cite{Mullenbach2018}. 
CAML models clinical coding as a multilabel classification problem -- given the input clinical document, output the set of billing codes. 
CAML proceeds by encoding patient notes with word vectors \cite{Mikolov2013DistributedRO}, then forming a contextualized representation by passing these to a convolutional neural network. 
CAML then uses a label-aware attention mechanism -- dot-product attention between label and textual representations -- to produce a single prediction per ICD10-CM code.


While this approach achieves competitive results in predicting ICD9 Codes for MIMIC-III, it comes with drawbacks. 
First, the codes in the training set determine the codes that CAML can predict, causing train-test mismatch issues. As shown in Table~\ref{tabl:code_mismatch}, 39.3\% (MIMIC-III) and and 27.4\% (Internal) of the test codes are not in the training set, and thus CAML can never predict these. 
Even predicting rare codes is challenging, as CAML needs to learn a dot-product attention for every code from scratch.


\subsection{Training}


Models are implemented with GluonNLP~\cite{guo2020gluoncv} (Base Model (BERT)) and Huggingface Transformers~\cite{wolf-etal-2020-transformers} (Longformer (codes, entities), Aggregation model). 
We trained our models on a hybrid dataset that combines MIMIC-III and our internal dataset. 
Hyperparameters (learning rate ($lr$), ranking loss interpolation ($\lambda$), and epochs) of the models in Section~\ref{sec:ranking} are selected via grid search using the combined validation set. We did not look at the test set before fixing all hyperparameters. 
Base Model (BERT) is trained for 20 epochs (30,027 entities per epoch) and has a $lr=1e-6$. 
Similarly, Base Model (BERT) + ranking loss has the same hyperparameters, with $\lambda=0.5$. 
Longformer (entities) is trained for 5 epochs (30,027 entities per epoch) with a $lr=1e-5$. 
Longformer (entities) + ranking loss is trained for 10 epochs with a $lr=1e-5$ and $\lambda=0.5$. 
Longformer (codes) is trained for 10 epochs (17,746 codes per epoch) with a $lr=1e-5$.
Longformer (codes) + ranking loss has the same hyperparameters with $\lambda=0.5$. 
Aggregation model is trained for 5 epochs (17,746 codes per epoch) with a $lr=1e-5$. 
Aggregation model + ranking loss is trained for 5 epochs with a $lr=5e-5$ and $\lambda=0.5$. 
Finally, CAML is trained for 1,553 epochs (567 documents per epoch) with a learning rate of 1e-4. During training of CAML, we found that the validation F1 remains zero until it reach around 700 epochs. This is partially due to the small number of sample size (567 documents) and very low prevalence rate (0.93\% relevant codes). For binary selection, prediction thresholds are chosen to maximize F1 on the combined validation set. For test set results reported in the paper, we report MIMIC-III and our internal dataset separately.

\subsection{Metrics}

We believe that code-ranking metrics are more important to evaluate system performance than are classification metrics. Ranking metrics directly measure properties relevant for a billing system -- how well does your model suggest codes to human annotators. As such, we include a discussion on ranking metrics suitable for billing systems.


\subsubsection{Ranking Metrics}
\paragraph{NDCG@12.} Discounted Cumulative Gain (DCG)\cite{jarvelin2002cumulated} is a graded relevance rank-based measure that models a determined user going down a ranked list. When a relevant item (in this case, ICD10-CM billing codes) is found by the user, a gain associated with the relevance of the found item is added to the score. However, the further down the ranked list, the less utility is the found item. Thus, a discount associated with the rank position is applied to the gain:
$$DCG(L) = \sum_{i=1}^{|L|} \frac{rel_{L_i}}{log_2(i+1)}$$
where $L$ is a ranked list and $rel_{L_i}$ is the relevance of the $i$-th item in ranked list $L$. The discount, $\frac{1}{log_2(i+1)}$ is a position-based discount that encourages finding relevant item closer to the top of the ranked list. Since DCG can be unbounded, a normalization term (Normalized DCG, thus NDCG) is often added to scale DCG between 0 and 1: $NDCG(L) = \frac{DCG(L)}{IDCG(L)}$, where $IDCG(L)$ is the maximum DCG score achievable by reordering $L$.

For our purpose of ranking and retrieving ICD10-CM codes from clinical documents, we set the relevance of the primary codes to be $2^3 - 1 =7$ and secondary codes to be $2^1 - 1 = 1$. To penalize the model for returning irrelevant codes, we additionally set the relevance of the irrelevant codes to be $-1$. Additionally, we focus on at most the first 12 items (codes) in the ranked list, conventionally denoted as NDCG@12 (see Section \ref{sec:data}).



\paragraph{P@1.} We report precision at 1, the percentage of time where the top-ranked code (according to the model) is the primary ICD10-CM code. 
Prior work reports precision cutoff at 5 and 8. However, we argue that these measures are not suitable for measuring ICD10-CM coding as they ignore relative code importance. 

\subsubsection{Classification Metrics}
\paragraph{Precision/Recall/F1} We calculate precision, recall, and F1 scores by comparing the model predictions with the ground truth annotation. We aggregate the scores three ways, each corresponding to different usages. 
\textit{Micro}: flatten all predictions and score the aggregate. This measures \textit{dataset-level} performance, and can be sensitive to outlier documents, with many or few codes. 
\textit{Doc} calculate precision, recall, and F1 score per document, then average across all documents. This represents the expected performance of the model at a document level. 
\textit{Code} calculate F1 score per ICD10-CM code, and then average across all codes appearing in the test set. Prior literature often reports this score as a \textit{Macro} average, and frequently restricts reporting to either the training set or the most frequent 50 codes.
Since we observe the large mismatch of codes between training, validation, and test set (see Table~\ref{tabl:code_mismatch}), we report our score over the test set codes to reflect the performance at test time.

\paragraph{ROC\_AUC.} Area Under the Receiver Operating Characteristic Curve (ROC\_AUC). ROC plots true positive rate and false positive rate over different prediction threshold. We calculate ROC\_AUC using the \textit{Micro} method, by flattening the predictions across documents.

\section{Results} \label{sec:results}

We report results on all methods we describe, including the additional baseline method of CAML. 
Table \ref{tabl:mimic_score} reports results on the MIMIC-III subset and our internal dataset.
We provide a detailed breakdown of classification results in Table \ref{tabl:classification_scores} by ICD10-CM code prevalence.

\begin{table}[t]

\small
\centering
\resizebox{\textwidth}{!}{\begin{tabular}{llrrrrrr}
\toprule
&{} &  NDCG@12 &   P@1 &           Micro &             Doc &            Code &  AUC \\
&Model                 &          &       &  Prec/Recall/F1  &  Prec/Recall/F1  &    Prec/Recall/F1   &          \\
\midrule
\parbox[t]{2mm}{\multirow{9}{*}{\rotatebox[origin=c]{90}{MIMIC-III}}} &CAML                  &    14.55 & 18.18 &  43.0/34.8/38.4 &  43.9/30.2/32.4 &  26.8/27.9/26.1 &    62.08 \\ \cmidrule{2-8}
&Base Model (BERT)      &    39.41 & 13.64 &  57.7/63.8/60.6 &  57.9/64.1/58.1 &  57.3/59.4/57.5 &    88.94 \\
&+ ranking loss        &    41.47 & 18.18 &  57.2/67.4/61.9 &  60.7/69.0/61.0 &  61.5/63.4/61.5 &    89.17 \\
&Longformer (Codes)    &    30.76 & 22.73 &  50.6/58.2/54.1 &  52.4/58.7/52.3 &  54.4/55.3/53.6 &    83.02 \\
&+ ranking loss        &    33.71 & 13.64 &  47.1/63.8/54.2 &  52.9/65.7/55.1 &  60.3/62.2/60.0 &    84.12 \\
&Longformer (Entities) &    51.53 & 31.82 &  59.7/65.2/62.4 &  64.5/71.3/63.5 &  65.0/64.1/63.6 &    88.75 \\
&+ ranking loss        &    51.51 & 40.91 &  58.5/68.1/63.0 &  61.0/73.4/62.8 &  67.2/66.4/65.6 &    88.94 \\
&Aggregation Model       &    \textbf{61.03} & 45.45 &  62.4/83.7/\textbf{71.5} &  66.6/85.9/\textbf{71.4} &  77.9/81.5/\textbf{78.6} &    \textbf{93.58} \\
&+ ranking loss        &    58.63 & \textbf{54.55} &  69.4/61.0/64.9 &  68.2/60.7/61.0 &  59.0/57.2/57.2 &    92.52 \\
\midrule \midrule
\parbox[t]{2mm}{\multirow{9}{*}{\rotatebox[origin=c]{90}{Internal Dataset}}} &CAML                  &    20.32 & 12.20 &  49.4/59.0/53.8 &  45.1/49.9/44.2 &  38.2/45.4/39.8 &    71.42 \\ \cmidrule{2-8}
&Base Model (BERT)      &    41.78 & 14.63 &  70.0/60.0/64.6 &  72.5/59.5/61.4 &  50.7/52.0/50.4 &    88.99 \\
&+ ranking loss        &    40.97 & 14.63 &  69.6/64.4/66.9 &  71.7/59.3/60.7 &  57.6/59.2/57.1 &    89.63 \\
&Longformer (Codes)    &    49.86 & 34.15 &  68.2/64.1/66.1 &  67.2/65.3/64.6 &  58.8/60.6/58.6 &    87.21 \\
&+ ranking loss        &    51.96 & 34.15 &  65.8/67.1/66.4 &  65.8/68.8/65.5 &  62.2/63.6/61.9 &    86.28 \\
&Longformer (Entities) &    62.45 & 35.37 &  78.0/73.2/75.5 &  80.5/76.8/76.1 &  68.4/68.6/67.6 &    90.97 \\
&+ ranking loss        &    64.11 & 41.46 &  76.1/73.6/74.8 &  75.2/78.2/74.4 &  71.2/71.2/\textbf{70.3} &    90.50 \\
&Aggregation Model       &    \textbf{68.22} & \textbf{46.34} &  77.9/75.3/\textbf{76.6} &  81.7/78.1/\textbf{77.7} &  69.6/70.4/69.2 &    \textbf{93.29} \\
&+ ranking loss        &    64.34 & 43.90 &  86.3/62.0/72.2 &  88.0/66.8/73.2 &  62.6/58.9/59.8 &    91.57 \\
\bottomrule
\end{tabular}
}
\caption{Results on MIMIC-III (top) and internal (bottom) test set. NDCG@12 is the ranking measure NDCG cutoff at 12. P@1 is the percentage of the highest predicted code being the primary billing code. \textit{Micro}, \textit{Doc}, \textit{Code} is the \textbf{precision}/\textbf{recall}/\textbf{F1} scores of different aggregation methods.}
\label{tabl:ezdi_score}\label{tabl:mimic_score}
\vspace{-1em}
\end{table}


\textbf{Classification Results}. We find that CAML performs poorly across all classification metrics and appears to be heavily data-limited due to the differences in our dataset size (567 documents) vs. the full MIMIC-III (47,724 documents).
In particular, Table \ref{tabl:classification_scores} shows that CAML, while showing competitive performance on the common codes, cannot classify unseen codes - an inherent limitation of the multi-label classification approach.
Increasing context integration greatly improves classification performance.
On both datasets, the Base Model substantially outperforms CAML, especially on rare and unseen codes (see Table~\ref{tabl:classification_scores}). This is partially enabled by the Base Model's ability to learn from ICD10-CM code description.
Of the two Longformer variations, entity-level relevance assignment outperforms code-level, suggesting that the context of the entities in the document is a better predictor than the code descriptions.
Finally, by jointly modeling the code description together with the context of the entities, the aggregation model performs the best of all models across almost all classification metrics.



\textbf{Ranking Results}. When the task of interest is \textit{ranking}-focused, rather than merely classification, our results show that careful integration of context can lead to large improvements.
For our internal dataset (Table \ref{tabl:ezdi_score}, bottom), moving from the Base Model to a Longformer-based approach to the aggregation model forms a clear trend that more context improves both absolute ranking NDCG and first place ranking P@1.
Trends in MIMIC-III results (Table \ref{tabl:mimic_score}, top) are similar, result improving with better integration of context. However, the Longformer code-level classifier performs uniformly worse than all other models (excepting the baseline).


\textbf{Ranking Loss}. Our most surprising result is that the ranking loss functions are not a universal improvement. Between datasets, it fails to consistently improve performance on any of the ranking metrics; occasionally having a large increase in one case, and the opposite effect in the next, e.g. MIMIC-III vs. Internal P@1 for the Entity-Based Longformer models (Table \ref{tabl:mimic_score}).

\textbf{Ranking vs. Classification}. Classification results can be misleading - despite high AUCs and decent F1s 
(Table \ref{tabl:mimic_score})
not all models perform as well on the ranking metrics. CAML performs well at overall classification, but performs poorly for  rare codes (Table \ref{tabl:classification_scores}), and its ranking scores are poor compared to other models. The models in this paper were developed targeting ranking performance, and perform well on all metrics, including rare and unseen codes.

\begin{table}[t]

\small
\centering
\resizebox{0.75\textwidth}{!}{
\begin{tabular}{lrrrrrr}
\toprule
{} & \multicolumn{3}{l}{Internal Dataset} & \multicolumn{3}{l}{MIMIC III} \\
{Model} &       Common &  Rare & Unseen &       Common &  Rare & Unseen \\
\midrule
CAML                  &        52.03 & 43.16 &   0.00 &        58.18 & 27.27 &   0.00 \\
Base Model (BERT)     &        56.53 & 48.84 &  53.33 &        62.23 & 59.54 &  47.83 \\
+ ranking loss        &        58.63 & 57.38 &  53.33 &        57.02 & 61.37 &  65.22 \\
Longformer (Entities) &        70.94 & 65.45 &  \textbf{78.33} &        57.22 & 63.05 &  69.57 \\
+ ranking loss        &        67.91 & 69.62 &  78.33 &        62.02 & 66.44 &  65.22 \\
Longformer (Codes)    &        47.25 & 61.33 &  53.33 &        50.88 & 54.61 &  52.17 \\
+ ranking loss        &        48.50 & 63.38 &  68.33 &        54.42 & 56.85 &  73.91 \\
Aggregation Model     &        \textbf{72.69} & \textbf{69.80} &  60.00 &        \textbf{68.90} & \textbf{80.81} &  \textbf{78.26} \\
+ ranking loss        &        65.14 & 60.19 &  50.00 &        66.82 & 55.28 &  56.52 \\
\bottomrule
\end{tabular}
}
\caption{F1 scores average across codes, broken down by two datasets and different code subsets. 
\textbf{Common} is the 50 most frequent codes in the training set. \textbf{Rare} is the codes not covered by \textbf{Common} in the training set.
\textbf{Unseen} is the codes not in the training set, but found in the test set (39.3\% of codes for MIMIC-III and 27.4\% of codes for internal test set are unseen in training).
\label{tabl:classification_scores}
}
\vspace{-1em}
\end{table}

\section{Conclusions} \label{sec:conclusions}

We have identified common issues in building and measuring automated medical coding systems, and proposed both models and methods to accommodate for those issues. 
Most modern systems for medical coding ignore result ordering. 
We have characterized metrics for evaluating these issues; our classification and ranking results show the importance of using appropriate ranking measures for building medical coding systems in practice.
Systems focusing on classification performance alone can produce misleading results and may result in poor model selection, and thus poor performance in the field, costing money and time.

Many existing approaches attempt to leverage the ICD code hierarchy, code descriptions, and other methods for forming code representations. 
Some of these approaches use a learned representation of the ICD10-CM code, and transform the coding task to a multi-label classification problem. 
Rather than limiting ourselves to a fixed set of codes, for which we may lack data, we have shown that the appropriate use of an ICD code's description can be sufficient for strong model performance and generalization to unseen codes.

Prior work in integrating transformers suffers from limitations in the maximum input size.
By building upon local contexts, anchored to medical entities, our models do not suffer from these limitations and are able to process documents regardless of input size. 
Our evaluations show that while existing approaches are helpful in addressing limitations inherent to transformer models for medical coding, they are not practical for assisting medical coders. 

Our results show that our approaches of building on local context, using ICD10-CM code descriptions, and proper task measurement can produce higher quality models for ICD-coding with less data than other systems require.
%

\bibliography{amia}  
\bibliographystyle{vancouver}

\end{document}